\documentclass{article}

\usepackage{arxiv}

\usepackage[utf8]{inputenc} 
\usepackage[T1]{fontenc}    
\usepackage[colorlinks=true, allcolors=blue]{hyperref}       
\usepackage{url}            
\usepackage{booktabs}       
\usepackage{amsfonts}       
\usepackage{nicefrac}       
\usepackage{microtype}      
\usepackage{lipsum}

\usepackage{graphicx}
\usepackage{graphics}
\usepackage{subfigure}

\usepackage{tikz}
\usetikzlibrary{shapes,arrows}
\tikzstyle{box} = [rectangle, text centered, draw=black]
\tikzstyle{oval} = [rounded rectangle, text centered, draw=black]
\tikzstyle{arrow} = [thick,->,>=stealth]
\newenvironment{diagram}{
\begin{equation}
\begin{tikzpicture}[node distance=1.5cm]
}
{\end{tikzpicture}
\end{equation}
}

\usepackage[colorinlistoftodos,disable]{todonotes}
\newcommand{\TODO}[1]{\todo[inline, color=green!40]{#1}}

\renewcommand{\TODO}{}
\usepackage{comment}
\title{A Boxology of Design Patterns for \\
Hybrid Learning and Reasoning Systems}

\author{
  Frank van Harmelen\\
  Department of Computer Science\\
  Vrije Universiteit Amsterdam\\
  \texttt{Frank.van.Harmelen@vu.nl} \\
   \And
  Annette ten Teije \\
  Department of Computer Science\\
  Vrije Universiteit Amsterdam\\
  \texttt{Annette.ten.Teije@vu.nl} \\
}

\begin{document}
\maketitle

\begin{abstract}
We propose a set of compositional design patterns to describe a large variety of systems that combine statistical techniques from machine learning with symbolic techniques from knowledge representation. As in other areas of computer science (knowledge engineering, software engineering, ontology engineering, process mining and others), such design patterns help to systematize the literature, clarify which combinations of techniques serve which purposes, and encourage re-use of software components. We have validated our set of compositional design patterns against a large body of recent literature. 
\end{abstract}


\section{Motivation}
\label{sec:motivation}

Recent years have seen a strong increase in interest in combining Machine Learning methods with Knowledge Representation methods. The interest in this is fuelled by the
complementary functionalities of both types of methods, and by their complementary strengths and weaknesses. This is witnessed by keynote addresses at all the major conferences in recent years (Hector Geffner  and Josh Tenenbaum at ECAI/IJCAI 2018 \cite{Geffner:IJCAI2018,Tenenbaum:2017}, Lise Getoor at NIPS 2017 \cite{Getoor:POD2017}, and William Cohen at ILP 2018 \cite{Cohen:NIPS2018} to name just a few in the last two years), and by publications that have attracted widespread attention such as \cite{DeepMind:2018} and \cite{Marcus:2018}. The workshop series on Neural-Symbolic Learning and Reasoning stretches back to over a decade\footnote{\url{http://neural-symbolic.org/}} but has seen a sharp rise in interest in recent years, and many of the major conferences have dedicated workshops to the topic, such as the MAKE workshop at the AAAI Spring Symposium 2018\footnote{\url{https://www.aaai-make.info/}}, the Induce and Deduce workshop\footnote{\url{https://sites.google.com/view/r2k2018/home}} and the workshop on Hybrid Reasoning and Learning\footnote{\url{https://www.hybrid-reasoning.org/kr2018_ws/}} at KR 2018, the workshop on Relational Representation Learning\footnote{\url{https://r2learning.github.io/}} at NIPS 2018 and many others. 

This increasing interest has resulted in an explosion of a large volume of diverse papers in a variety of venues, and from a variety of communities (of course from machine learning and knowledge representation, but also from semantic web, from natural language, from cognitive science, etc). Both this volume and this diversity of origin has created a very diffuse literature on the subject, with no consensus of which approaches are promising, using very different formalisms (ranging from graph theory to linear algebra and continuous differentiable functions), different architectures, different algorithms,often even different vocabularies to speak about the same concepts depending on the community of origin, and spread out over a large space of journals and conference, typically not surveyed by any single researcher. 

This paper is an attempt to create some structure in this large, diverse and rapidly growing literature. We do not think it is possible anymore to provide an exhaustive overview of the entire literature. The literature is simply too voluminous, heterogeneous and spread out for such an exhaustive overview. Instead, what we aim for in this paper is to present a conceptual framework that can be used to categorize much if not all of the techniques for combining learning and reasoning. Our framework takes the form of a set of design patterns, which we will motivate in the next section. Our claim is \emph{not} the that we cite and discuss the complete relevant literature on this topic (as stated, we think this would be impossible by now). Instead, our completeness claim is that our set of  patterns covers all design variations that appear in the literature. Thus, referring to an additional paper would not by itself be an extension to this work, but would only be an extension to this work if such a paper describes a hybrid learning-and-reasoning architecture not yet covered by our set of patterns.

We have validated our set of design patterns against a set of more than 50 papers from the research literature from the last decade. 
Our claim is that each of the systems that we encountered in those references is captured by one of our design patterns. 

In the next section we will discuss the basic distinction between learning systems and reason systems. In section 3 we introduce our graphical notation, before the main section 4 where we present our library of compositional patterns. Section 5 discusses future work and section 6 concludes. 

\section{The two families of techniques}
\label{sec:two-families}

Although not universal, there is some consensus in the literature as to the need for combining methods for learning (which are most predominantly statistical in nature) with methods for reasoning (which are predominantly discrete in nature): ``Our general conclusion is that human-level AI cannot emerge solely from model-blind learning machines; it requires the symbiotic collaboration of data and models''
 \cite{PearlImpediments:2018}; ``By pushing beyond perceptual classification and into a broader integration of inference and knowledge, artificial intelligence will advance greatly.'' \cite{Marcus:2018} ; ``the question is not whether it is functions or models but how to profoundly integrate and fuse functions with models''\footnote{See our discussion for an explanation of this terminology.} \cite{Darwiche:CACM2018}.
 
 Other papers extensively discuss the advantages and disadvantages of both types of approaches in depth, but we briefly summarize the main points here, citing from \cite{Marcus:2018} and the introductory section of \cite{DeepSymbolicReinforcementLearning:2016}:

Limitations of (deep) learning systems:
\begin{itemize}
\item \textbf{Data hungry:} Learning systems (and in particular modern deep learning systems) need very large training sets; 
\item \textbf{Limited transfer:} a trained network that performs well on one task often performs very poorly on a new task, even if the new task is very similar to the one it was originally trained on; 
\item \textbf{Brittle:}  they are susceptible to adversarial attacks, meaning that even for seemingly similar inputs for the same task, the outputs may differ significantly; 
\item \textbf{Opaque:} and they are opaque, meaning that is typically difficult to extract a humanly-comprehensible chain of reasons for output of the system. 
\item \textbf{No use of prior knowledge:} the performance of learning systems is based on the data they see during their training phase, and is not informed by general principles such as causality, or general domain knowledge.
\end{itemize}

Limitations of symbolic reasoning systems:
\begin{itemize}
\item \textbf{Brittle:} their foundation in discrete formalisms makes it hard to capture exceptions, and make these systems very unstable in the presence of noisy data. 
\item \textbf{Size:} acquiring explicit knowledge-bases (typically from experts) is error-prone and expensive, typically limiting the scope of such systems
\item \textbf{Efficiency:} the logic-based reasoning methods are typically subject to combinatorial explosions that limit both the number of axioms, the number of individuals and relations described by these axioms, and the depth of reasoning that is possible. 
\end{itemize}

Many authors in the literature have noted that these two sets of limitations are strongly complementary, and that furthermore the two families of techniques have seen successes in very different application scenario's, with statistical learning techniques successful in pattern recognition (e.g. image interpretation, speech recognition, natural language translation, board games and video games), while the successes of symbolic reasoning techniques are in such applications as planning (e.g. in robotics), diagnosis, design tasks and question answering (e.g. in personal assistants).  This begs the question of the more precise delineation of these two families of systems. 

\subsubsection*{Task-based distinction}

A first characterization found in the literature is based on the \emph{task} performed by the system:

\textbf{Deduction vs. Induction}: the classical distinction between reasoning and learning is the late 19th century Peirce-ian distinction between deduction and induction. Deduction derives specific conclusions from general statements, where conversely induction derives general statements from specific observations. More formally, deduction is the derivation of specific conclusions $\phi$ given a set of general formulae $T$ and a deductive calculus $\vdash$: $T \vdash \phi$. Conversely, induction is the derivation of a set of general formulae $T$ given a set of specific observations $\phi_1,\ldots,\phi_n$ such that $T \vdash \phi_i$ for all $i=1,\ldots,n$: deduction is the problem of deriving $\phi$ given $T$, while induction is the problem of deriving $T$ given $\phi_1,\ldots,\phi_n$.\footnote{Confusingly, in logic $T$ is called a ``theory'', while in machine learning, $T$ would be called the ``model'', whereas the term ``model'' is used in a very different sense in logic.} 

\textbf{Compression vs. decompression}: a more recent characterization is that of learning as compression \cite{Vitanyi:2005}. The intuition here is that an inductive learning process ``compresses'' a large set of observations $\phi_1,\ldots,\phi_n$ into a more compact model $T$, using the Minimum Description Length principle \cite{MinDescrLength:1998} as a complexity measure for both data and model. Conversely, reasoning is then the process of producing (other) predictions $\phi$ from such a model $T$, which can be seen as a form of decompression: after all, all the conclusions $\phi$ were already ``implicitly'' present in $T$, and the job of deduction is to simply ``decompress'' the general theory $T$ into the more specific conclusions $\phi$. 

Both of these characterizations impose a strict dichotomy between the two modes of learning and reasoning. Other authors have instead tried to conceptualize a continuum of options that interpolate between induction and deduction (e.g. \cite{continuum:2011}), but these have not been widely adopted. 

\subsubsection*{Representation-based distinction}

Whereas the above dichotomies tried to capture the different \emph{tasks} that are performed by reasoning or learning systems, another popular and somewhat orthogonal distinction in the literature is based on the \emph{representation} that is used by different systems.

Pearl \cite{PearlImpediments:2018} uses the term ``model-free'' for the representations typically used in many learning systems, and Darwiche \cite{Darwiche:CACM2018} described them as ``function-based''\footnote{Other names used for inferences at this layer are: “model-blind,” “black-box,” or “data-centric” \cite{PearlImpediments:2018}}, to emphasize that the main task performed by deep learning systems is function-fitting: fitting data by a complex function defined by a neural network architecture. Such ``function-based'' or ``model-free'' representations are in contrast to the ``model-based'' representations typically used in reasoning systems. 

There is no consensus in the literature what precisely constitutes such a ``model-based'' representation, but typical properties that are ascribed to such model-based representations are that they are 
\begin{itemize}
\item \textbf{compositional:} the meaning of model is a function of the meaning of its components, 
\item \textbf{referential:} the model is constructed out of symbols that refer to objects and relations in the world
\item \textbf{homologous:} the structure of the model mirrors the structure of the world it is modelling,  
\item \textbf{interpretable:} the structure and content of a model is human-understandable and traceable, 
\item \textbf{symbolic:} as opposed to numeric, 
\item \textbf{discrete:} as opposed to real-valued, continuous and differentiable. 
\end{itemize}

Examples of such models are of course logical formalisms (such as propositional, first-order, modal and non-monotonic logics \cite{KRHandbook}), but also grammars, knowledge graphs \cite{knowledgeGraphs:2016}, ontologies \cite{OntologiesHandbook}, graphical models \cite{KollerBook}, models from qualitative physics \cite{StrussQR}, etc. 

Notice that the distinction of ``model-based'' vs. ``model-free'' (or: ``function-based'') representations is orthogonal to the previously discussed task-based dichotomy: systems such as Markov Logic Networks \cite{MarkovLoginNetworks} do perform a learning task (in order to learn weights on the relations between variables from data), but do so based on an explicit (graphical) model. Similarly, Inductive Logic Programming (ILP \cite{ILPSpecialIssue:2017}) performs a learning task, but again does so on an explicit model (and even on a model (Horn Clauses) originally intended for deductive reasoning). In the words of Darwiche \cite{Darwiche:CACM2018}: \emph{``Machine learning [...] has a wide enough span that it overlaps with the model-based approach; for example, one can learn the parameters and structure of a model but may still need non-trivial reasoning to obtain answers from the learned model''}. 
Nevertheless, it would be fair to say that the vast majority of modern work on machine learning (and in particular work on deep learning) uses a ``model-free'' (or: ``function-based'') representation of data. 

We will use both of the above distinctions (task-based and representation based) in the design patterns that we will introduce next. 

\section{Design Patterns and notation}

The notion of re-usable design patterns has been successfully used in many different areas of Computer Science. Perhaps the best known of these are the Design Patterns from Software Engineering \cite{DesignPatterns:1994,DesignPatterns:2010}. These Design patterns have successfully captured general reusable solutions to commonly occurring problems in software design in the form of a template for how to solve a problem, that can be used in many different situations. These design patterns are organized in a hierarchical taxonomy, and typically expressed in a graphical notation\footnote{sometimes described tongue-in-cheek as a \textit{boxology}: ``A representation of an organized structure as a graph of labelled nodes and connections between them'', \url{https://www.definitions.net/definition/boxology}}.
Around the same time, a similar set of design patterns were developed for Knowledge Engineering in the form of the CommonKADS task library \cite{CommonKADS:1994,commonKADS:1999}. In a similar vein to the patterns from Software Engineering, the CommonKADS library identified frequently occurring problem templates (called ``tasks''), together with known templates for solving these problems (called ``inference structures''), these templates were again organized in a hierarchical taxonomy, and expressed in a UML-like graphical notation. Other examples of Computer Science subdisciplines which have developed such design patterns are ontology design \cite{Gangemi:2009} and process mining \cite{processmining}. 

Broadly recognized advantages of such design patterns are  they distill previous experience in a reusable form for future design activities, they encourage re-use of code, they allow composition of such patterns into more complex systems, they provide a common language in a community, 
and they are a useful didactic device \cite{DesignPatternsAdvantages:1998,DesignPatternsAdvantages:2001}. 
In this paper we aim to define such a set of design patterns for capturing the wide variety of theories, proposals and systems in the literature on hybrid systems that combine learning and reasoning. 
Our patterns distinguish between systems both on the functionality of their components and on the representations that they deploy, using both of the distinctions discussed above. 
We show that our patterns are indeed compositional (complex configurations can be built by composing simple architectures), and we claim a substantial degree of completeness for our library of compositional patterns,
by validating them against a body of more than 50 papers from the research literature of the past decade. 

\label{sec:notation}

We will now introduce the informal graphical ``boxology'' notation that we use to express our patterns. We  use ovals to denote algorithmic components (i.e. objects that perform some computation), and boxes to denote their input and output (i.e. data structures). Following the task-based dichotomy described above, we distinguish two types of algorithmic components (ovals): those that perform some form of deductive inference (labelled as the ``KR'' components) and those that perform some form inductive inference (the ``ML'' components):
\begin{tikzpicture}
\node (KR) [oval] {KR};
\node (ML) [oval, right of=KR] {ML}; 
\end{tikzpicture}~.
Based on the representation-based distinction discussed above, we also use two kinds of input- and output-boxes: those that contain ``model-based'' (symbolic, relational) structures, those that contain ``model-free'' data:
\begin{tikzpicture}
\node (sym) [box] {sym};
\node (data) [box, right of=KR] {data}; 
\end{tikzpicture}~.

\noindent
The \emph{sym}-boxes are the input and output of a classical KR
reasoning system:

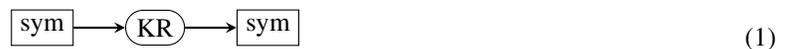
\begin{diagram}
\node (in) [box] {sym};
\node (KR) [oval, right of=in] {KR};
\draw [arrow] (in) -- (KR) ;
\node (out) [box, right of=KR] {sym};
\draw [arrow] (KR) -- (out) ;
\label{pattern:simple-KR}
\end{diagram}

\noindent
and idem the data boxes are the typical input and output boxes of an ML system:

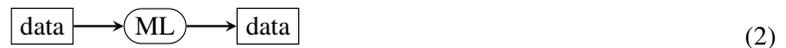
\begin{diagram}
\node (in) [box] {data};
\node (ML) [oval, right of=in] {ML};
\draw [arrow] (in) -- (ML) ;
\node (out) [box, right of=ML] {data};
\draw [arrow] (ML) -- (out) ;
\label{pattern:simple-ML}
\end{diagram}

Based on the discussion in the previous section, the labels ``inductive'' and ``deductive'' for the algorithmic components would have been more accurate, but we use the labels ``KR'' and ``ML'' for brevity. Similarly, the labels ``model-based'' and ``model-free'' (or ``model-based'' and ``function-based'') would have been more accurate for the input- and output-boxes, but again we use the labels ``sym'' and ``data'' for brevity. 


\section{A library of Patterns}
\label{sec:patterns}

In this section we identify common patterns for hybrid systems that perform reasoning and learning. 

\subsection*{Learning with symbolic input and output}

\noindent
Instead of applying ML techniques to model-free data such as images, text or numbers, the ML
techniques can be applied to symbolic structures, also yielding symbolic output:

\begin{diagram}
\node (in) [box] {sym};
\node (ML) [oval, right of=in] {ML};
\draw [arrow] (in) -- (ML) ;
\node (out) [box, right of=ML] {sym};
\draw [arrow] (ML) -- (out) ;
\label{pattern:ILP}
\end{diagram}

A classical examples of this are the aforementioned approaches based on Inductive Logic Programming \cite{ILPSpecialIssue:2017,ProbILP:2014}, Probabilistic Soft Logic \cite{PSL:2013,PSL:2017} and Markov Logic Networks \cite{MarkovLoginNetworks}. Even this simple example shows the value of these abstract patterns: even though the algorithms and representations of ILP, PSL and MLN's are completely different, the design patterns shows that they are all aimed at the same goal: inductive reasoning over symbolic structures. 
\TODO{give different annotations of pattern (\ref{pattern:ILP})?}

\subsection*{From symbols to data and back again}
A more recent class of this ``graph completion'' systems \cite{Paulheim:JWS2017,Wang:2017,Nickel:2016} also satisfies this design pattern: a machine learning algorithm takes a knowledge graph as input and uses inductive reasoning to predict addition edges which are deemed to be true based on observed patterns in the graph, even though they are missing from the original graph. However, allmost all graph completion algorithms perform this task by first translating the knowledge graph to a representation in a high-dimensional vector space (a process called ``embedding''), to the following refinement of pattern (\ref{pattern:ILP}) would be more accurate: 

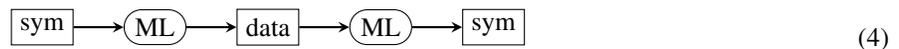
\begin{diagram}
\node (in) [box] {sym};
\node (ML1) [oval, right of=in] {ML};
  \draw [arrow] (in) -- (ML1) ;
\node (middle) [box, right of=ML1] {data};
  \draw [arrow] (ML1) -- (middle) ;
\node (ML2) [oval, right of=middle] {ML};
  \draw [arrow] (middle) -- (ML2) ;
\node (out) [box, right of=ML2] {sym};
  \draw [arrow] (ML2) -- (out) ;
  \label{pattern:embedding}
\end{diagram}
\TODO{annotate the boxes with knowledge-graph, vector-space, knowledge graph?}


\subsection*{Learning from data with symbolic output}
A variation of the above is when ML techniques are applied to model-free, but still yielding symbolic, model-based output: 

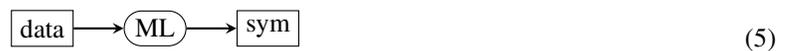
\begin{diagram}
\node (in) [box] {data};
\node (ML) [oval, right of=in] {ML};
\draw [arrow] (in) -- (ML) ;
\node (out) [box, right of=ML] {sym};
\draw [arrow] (ML) -- (out) ;  
\label{pattern:ontology-learning}
\end{diagram}

\noindent 
The typical example here is ontology learning from text \cite{OntologLearning}. Again, a large number of different approaches are captured by this single pattern: ontology learning using Inductive Logic Programming \cite{ILPOntologyLearning}, using conceptual spaces \cite{learningconceptualspaces} or text mining \cite{textminingOntologyLearning} are all described by pattern (\ref{pattern:ontology-learning}). A related but different instantiation of this patter is the use of text-mining not to learn full-blown ontologies, but to learn just the class/instance distinction (which is always problematic in ontology modelling), as done in
\cite{Patel-Schneider:EKAW2018}. As concerns the design patterns, this work only differs in the actual content of the symbolic output: a full-blown ontology, or only a class/instance label. 
\TODO{Produce two different annotated versions, one with text to ontology, and one with text to class/instance distinction?}

An entirely different application of this pattern is not to learn an ontology, but instead to learn a knowledge graph, as done in \cite{PSLLearningKnowledgeGraphs} by using Probabilistic Soft Logic as the learning engine. 
\TODO{again produce an annotated version of pattern \ref{pattern:ontology-learning}}
It is useful to note that many ``classical'' learning algorithms such as decision tree learning and rule mining are also covered by this design pattern.  

\subsection*{Explainable learning systems}

A major motivation for pattern (\ref{pattern:ontology-learning}) is the opaqueness problem mentioned in section \ref{sec:two-families}): the symbolic output is more amenable to crafting an explanation of the learning results \cite{WeldIntelligibleAI}. A natural extension of this pattern is therefore to use the symbolic output as input for a classical reasoning system, where the reasoning systems is used to craft an intelligible explanation of the results of the machine learner. 

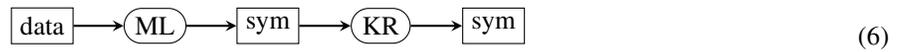
\begin{diagram}
\node (in) [box] {data};
\node (ML1) [oval, right of=in] {ML};
  \draw [arrow] (in) -- (ML1) ;
\node (middle) [box, right of=ML1] {sym};
  \draw [arrow] (ML1) -- (middle) ;
\node (KR) [oval, right of=middle] {KR};
  \draw [arrow] (middle) -- (KR) ;
\node (out) [box, right of=KR] {sym};
  \draw [arrow] (KR) -- (out) ;
  \label{pattern:simple-explanation}
\end{diagram}

\subsection*{Explainable learning systems with background knowledge}
An extension of this pattern describes the work in both \cite{Hitzler-Explaining:2017} and \cite{Tiddi-Explaining}, where background knowledge is used in the process of deductively reconstructing an explanation for the results of the learner: 

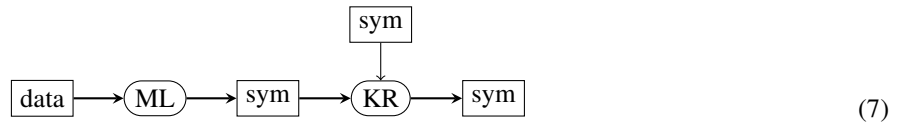
\begin{diagram}
\node (in) [box] {data};
\node (ML1) [oval, right of=in] {ML};
  \draw [arrow] (in) -- (ML1) ;
\node (middle) [box, right of=ML1] {sym};
  \draw [arrow] (ML1) -- (middle) ;
\node (KR) [oval, right of=middle] {KR};
  \draw [arrow] (middle) -- (KR) ;
\node (out) [box, right of=KR] {sym};
  \draw [arrow] (KR) -- (out) ;
\node (BG) [box,  above = 0.5cm of KR] {sym};
  \draw [->] (BG) -- (KR) ;
\end{diagram}

\subsection*{Explainable learning systems through inspection}
An alternative approach to explainable systems is taken in \cite{Pan-Explaining-Transfer-Learning}, where 
the behaviour of machine learning system (in this case: a neural net classifier trained with transfer learning)
is inspected by a reasoning system (in this case: a Description Logic reasoner), which then tries to explain the behaviour of the learner (in this case: which features were succesfully used in the transfer learning process). 

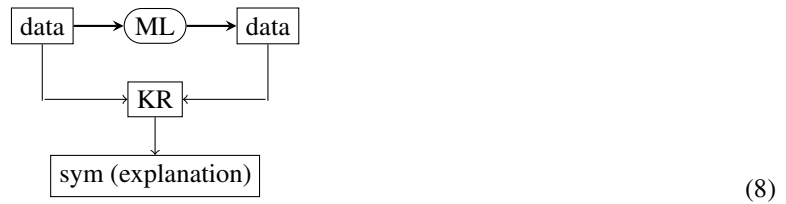
\begin{diagram}
\node (in) [box] {data};
\node (ML) [oval, right of=in] {ML};
  \draw [arrow] (in) -- (ML) ;
\node (out) [box, right of=ML] {data};
  \draw [arrow] (ML) -- (out) ;
\node (KR) [box, below = 0.5 cm of ML] {KR} ;  
\node (BG) [box,  below = 0.5cm of KR] {sym (explanation)};
  \draw [<-] (BG) -- (KR) ;
  
\draw [-] (in.south) -- ++(0,-0.75) ;
\draw [<-] (KR.west) -- ++(-1.1,0) ;

\draw [<-] (KR.east) -- ++(1.1,0) ;
\draw [-] (out.south) -- ++(0,-0.75) ;
\end{diagram}

\subsection*{Learning an intermediate abstraction for learning}

In pattern (\ref{pattern:embedding}) we have already seen a case where an intermediate representation is produced by one learning system as the input for a subsequent learning system. This turns out to be a rather generic pattern, of which also other variations are possible. One such variation is described in \cite{DeepSymbolicReinforcementLearning:2016}, where perceptual (``model-free'') input is used to learn an intermediate symbolic (``model-based'') representation of a the environment, and this symbolic spatial representation is then used in a reinforcement learning step to learn optimal behaviour:

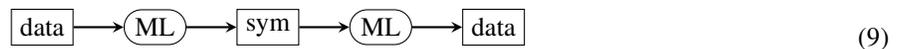
\begin{diagram}
\node (in) [box] {data};
\node (ML1) [oval, right of=in] {ML};
  \draw [arrow] (in) -- (ML1) ;
\node (middle) [box, right of=ML1] {sym};
  \draw [arrow] (ML1) -- (middle) ;
\node (ML2) [oval, right of=middle] {ML};
  \draw [arrow] (middle) -- (ML2) ;
\node (out) [box, right of=ML2] {data};
  \draw [arrow] (ML2) -- (out) ;
  \label{pattern:intermediate-spatial}
\end{diagram}
The results in \cite{DeepSymbolicReinforcementLearning:2016} show that the intermediate (and more abstract) symbolic representation gives a more robust behaviour of the system and allows for transfer learning between situations.
Besides learning a spatial abstraction (as in \cite{DeepSymbolicReinforcementLearning:2016}), the work in \cite{MultipleRLtasks} uses the same design pattern for deriving a temporal abstraction of  sequence of subtasks, which are then input to reinforcement learning agents.  
\TODO{Give two annotations of this pattern, for temporal and spatial?}

\subsection*{Learning an intermediate abstraction for reasoning}

Contrary to widespread popular belief, the Alpha Go system is not a single machine learning system. It is in fact built out of a machine learning component which learns functions for board valuation and region selection, which are subsequently used as components in a (classical) Monte Carlo search. This architecture can be described as 

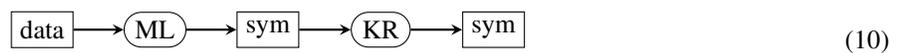
\begin{diagram}
\node (in) [box] {data};
\node (ML1) [oval, right of=in] {ML};
  \draw [arrow] (in) -- (ML1) ;
\node (middle) [box, right of=ML1] {sym};
  \draw [arrow] (ML1) -- (middle) ;
\node (KR) [oval, right of=middle] {KR};
  \draw [arrow] (middle) -- (KR) ;
\node (out) [box, right of=KR] {sym};
  \draw [arrow] (KR) -- (out) ;
\end{diagram}

\subsection*{Deriving an intermediate abstraction for reasoning}

In \cite{Hoogendoorn:2016} a raw data-stream is first abstracted into a stream of symbols with the help of a symbolic ontology, and this stream of symbols is then fed into a classifier (which performs better on the symbolic data than on the original raw data).

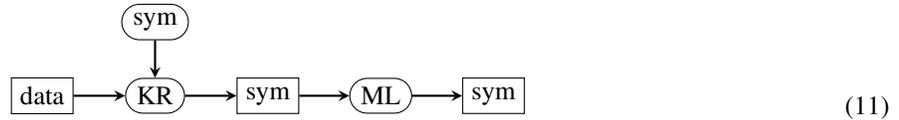
\begin{diagram}[node distance=1.5cm and 0.5cm]
\node (in) [box] {data};
\node (KR) [oval, right of=in] {KR};
  \draw [arrow] (in) -- (KR) ;
\node (onto) [oval, above = 0.5cm of KR] {sym};
\draw [arrow] (onto) -- (KR) ;
\node (middle) [box, right of=KR] {sym};
  \draw [arrow] (KR) -- (middle) ;
\node (ML2) [oval, right of=middle] {ML};
  \draw [arrow] (middle) -- (ML2) ;
\node (out) [box, right of=ML2] {sym};
  \draw [arrow] (ML2) -- (out) ;
\end{diagram}



%
%
%
%
\TODO{two patterns commented out here...}

\subsection*{Learning with symbolic information as a prior}

The following design patterns aims to resolve one of the issues mentioned in section \ref{sec:two-families}), namely how to enable machine learning systems to use prior knowledge:

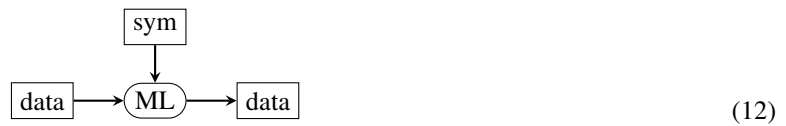
\begin{diagram}
\node (in) [box] {data};
\node (ML) [oval, right of=in] {ML};
\draw [arrow] (in) -- (ML) ;
\node (out) [box, right of=ML] {data};
\draw [arrow] (ML) -- (out) ;
\node (concl) [box, above = 0.5cm of ML] {sym};
\draw [arrow] (concl) -- (ML) ;
\label{pattern:inductive-bias}
\end{diagram}

An example of this are the Logic Tensor Networks in \cite{LTN:IJCAI2017}, where the authors show that encoding prior knowledge in symbolic form allows for better learning results on fewer training data, as well as more robustness against noise. A similar example is given in \cite{Tresp:ISWC2017}, where knowledge graphs are successfully used as priors in a scene description task, and in \cite{injecting-rules-into-embeddings} where logical rules are used as background knowledge for a gradient descent learning task in a high-dimensional real-valued vector space. 

The work by \cite{semantic-loss-function} is at first sight apparently unrelated: it investigates the use of a semantically formulated loss-function to drive the gradient descent learning process
(the semantic loss function is defined
as a propositional formula in conjunctive normal form, which is then made differentiable by weakening satisfiability to maximal satisfiability).
But when drawing the design pattern for this work we arrive at precisely diagram (\ref{pattern:inductive-bias}) above. This suggests that we could look at \cite{LTN:IJCAI2017} and \cite{Tresp:ISWC2017} with entirely different eyes, namely that they are in essence using their background knowledge as encoding a ``semantic loss'' function, and on closer inspection this is in fact a rather faithful account of what these papers are doing. This analogy was not mentioned at all in these papers, but was revealed by the design pattern that describes these systems. 

\subsection*{Learning with derived symbolic information as a prior}

Of course the ``inductive bias'' (using the terminology from \cite{DeepMind:2018}) does not need to be given, but can itself be derived by a reasoner, leading to a variation of pattern (\ref{pattern:inductive-bias}):

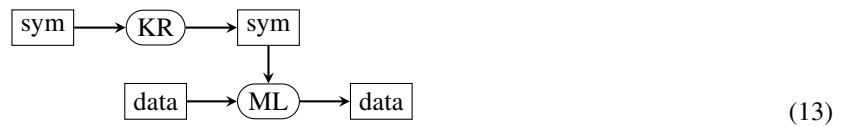
\begin{diagram}
\node (in) [box] {data};
\node (ML) [oval, right of=in] {ML};
\draw [arrow] (in) -- (ML) ;
\node (out) [box, right of=ML] {data};
\draw [arrow] (ML) -- (out) ;
\node (concl) [box, above = 0.5cm of ML] {sym};
\draw [arrow] (concl) -- (ML) ;
\node (KR) [oval, left of=concl] {KR};
\draw [arrow] (KR) -- (concl) ;
\node (prem) [box, left of=KR] {sym} ;
\draw [arrow] (prem) -- (KR) ;
\label{pattern:derived-prior}
\end{diagram}

\subsection*{Meta-reasoning for control}

There is a long-standing tradition in both AI \cite{meta-reasoning-survey} and in the field of cognitive architectures (e.g. \cite{PerlisBKM17}) to investigate so-called meta-reasoning systems, where one system reasons about (or:learns from) the behaviour of another system. 

In one pattern, also known as meta-cognition, symbolic reasoning is used to control the behaviour of a learning agent, to decide what it should learn and when, when it should stop learning, and in general to decide on the hyper-parameters that control the learning process:

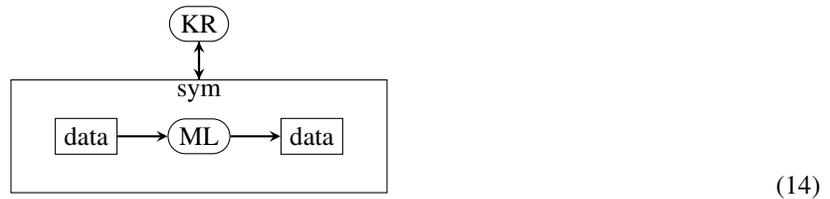
\begin{diagram}
\node (bb) [box, minimum height=1.5cm, minimum width=5cm] {};
\node [oval] at (bb.center) (ML) {ML};
\node (in) [box, left of=ML] {data};
\draw [arrow] (in) -- (ML) ;
\node (out) [box, right of=ML] {data};
\draw [arrow] (ML) -- (out) ;
\node (KR) [oval, above = 0.5cm of bb] {KR};
\draw [arrow] (KR) -- (bb)  ;
\draw [arrow] (bb) -- (KR) ;
\node (meta) [above = 0.1 cm of ML] {sym} ;
\end{diagram}

Here the KR system has a symbolic representation of the state of the ML system, reasons about it, and effectuates its conclusions as control instructions to the ML system. In a loose cognitive analogy, this could be compared with a consciously learning student, who constantly reflects on her learning progress to adjust her learning behaviour. 

\subsection*{Meta-reasoning for learning to reason}

In a second meta-reasoning pattern, the behaviour of one system (a symbolic reasoner) is the input of a second, machine learning, system. The machine learning system observes the behaviour of the symbolic reasoner, and learns from this behaviour how to perform deductive behaviour, which it is then able to mimic on new symbolic queries: 
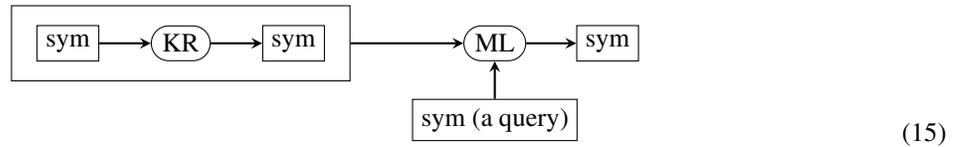
\begin{diagram}
\node (bb) [box, minimum height=1cm, minimum width=4.5cm] {};
\node [oval] at (bb.center) (KR) {KR};
\node (in) [box, left of=KR] {sym};
\draw [arrow] (in) -- (KR) ;
\node (outKR) [box, right of=KR] {sym};
\draw [arrow] (KR) -- (outKR) ;
\node (ML) [oval, right=of bb] {ML} ; 
\node (query) [box, below = 0.5cm of ML] {sym (a query)}; 
\draw [arrow] (bb) -- (ML) ;
\draw [arrow] (query) -- (ML) ;
\node (outML) [box, right of=ML] {sym} ;
\draw [arrow] (ML) -- (outML) ;
\end{diagram}
This pattern for training a neural network to do logical reasoning captures a wide variety of approaches such as reasoning over RDF knowledge bases \cite{Hitzler-learning-to-reason}, Description Logic Reasoning \cite{Lukasiewicz-learning-to-reason} and logic programming \cite{differentiable-proving}. 

\subsection*{Compositional systems}

The first of our patterns (patterns (\ref{pattern:simple-KR}), (\ref{pattern:simple-ML}), (\ref{pattern:ILP}) and (\ref{pattern:ontology-learning})) are the elementary building blocks out of which the more complex patterns can all be constructed. Compositionality can also be seen between more complex patterns, as in the meta-reasoning diagrams. 

For example, (\ref{pattern:embedding}) is a sequential  composition of (\ref{pattern:ILP}) and (\ref{pattern:ontology-learning}); pattern (\ref{pattern:derived-prior}) is a non-sequential of the two elementary patterns (\ref{pattern:simple-KR}) and (\ref{pattern:simple-ML}).

\section{Future work}

In future work, we intend to design a set of grammar rules that generate the space of all syntactically possible combinations of our atomic patterns. This should be followed by an attempt to find examples of all of these patters in the literature, and (if no exemplars are found in the literature) to investigate if these are meaningful combinations that have not yet been explored. 

A simple but interesting future extension of our notation would be to introduce a third type of processor (besides symbolic reasoner and learning system) namely that of a human agent. This would then allow our design patterns to be extended to human-in-the-loop systems, including the recently emerging family of hybrid intelligence systems that combine AI systems with humans in a single team. 

A more fundamental and almost philosophical issue to be addressed in future work is the distinction between \textit{data} and \textit{symbols}. Even though there is a shared intuition about this distinction (as in the earlier cited papers of Pearl \cite{PearlImpediments:2018} and Darwiche \cite{Darwiche:CACM2018}, we have not been able to come up with a crisp distinction, let alone a way to capture this distinction formally.

\section{Concluding comments}
\label{sec:conclusion}

What is notably different in our approach from other survey work in the literature is that we are not categorizing work based on the specific techniques that are being used inside the building blocks, but only on how the building blocks fit together. 
Each category abstracts from specific mathematical and algorithmic details of the specific approaches in that
category, but only looks at the functional behaviour of the pattern and at the functional dependencies between the ML and KR components.
This makes our categorization of systems in design patterns much more abstract and general. For example, while major battles are being fought in the literature between different forms of statistical relational learning, we abstract all of these approaches into a single design pattern (in this case pattern (\ref{pattern:ontology-learning}), allowing us to see that any of them could be deployed in more complex configurations such as (\ref{pattern:intermediate-spatial}) or (\ref{pattern:simple-explanation}). 

Based on the experience with design patterns in other subdisciplines of Computer Science such as Software Engineering and Knowledge Engineering, we hope that this classification of a wide variety of systems in a small number of compositional patterns will help with a better understanding of the design space of such systems, including a better understanding of the advantages and disadvantages of the different configurations, and a better understanding which design patterns are more suited for which kinds of performance tasks. Examples of this in the above were the use of an intermediate symbolic representation of space in \cite{DeepSymbolicReinforcementLearning:2016} to obtain more efficient and robust learning, the use of a symbolic representation in \cite{Hitzler-Explaining:2017} to produce explanations of the results of a classical learner, the use in \cite{Hoogendoorn:2016} of a symbolic reasoner to obtain a data abstraction which improved the performance of a subsequent learning algorithm, etc. 

Finally, we are experimenting with this approach as a didactic device\footnote{and in fact, two years of teaching a research seminar on combining statistical and symbolic approaches in AI has been our main motivator for this work.}. 

Although we have refrained from linking our design patterns to the design of cognitive architectures (see \cite{cognitive-architectures-survey} for a survey, it is tempting to do so. System such as ACT-R, SOAR, Sigma and others  distinguish components for temporal and spatial abstraction, for short and long term memory, for goal formulation and attention guidance, etc. Some of the patterns we have discussed are clearly reminiscent of some of these cognitive functions, and a study of these analogies would yield potentially interesting insights. 

Further obvious next steps in this work would be to perform a deeper analysis in which to apply these patterns to a wider body of literature, to formalize and further refine the informal descriptions in this paper, and to ultimately use this approach in a prescriptive design theory of statistical-symbolic systems.

\bibliographystyle{unsrt}  
\bibliography{boxology}

\end{document}